\documentclass[sigconf]{acmart}

\AtBeginDocument{%
  \providecommand\BibTeX{{%
    \normalfont B\kern-0.5em{\scshape i\kern-0.25em b}\kern-0.8em\TeX}}}

\setcopyright{acmlicensed}
\copyrightyear{2018}
\acmYear{2018}
\acmDOI{XXXXXXX.XXXXXXX}

\acmISBN{978-1-4503-XXXX-X/18/06}

\usepackage{enumitem}
\usepackage{multirow}
\usepackage{graphicx}
\usepackage{booktabs}
\usepackage{graphicx}
\usepackage{adjustbox}
\usepackage{cleveref}
\crefname{table}{tab.}{tabs.}
\usepackage{colortbl}
\makeatletter

\newcommand{\Rmnum}[1]{\expandafter\@slowromancap\romannumeral #1@}
\makeatother

\usepackage{subcaption}
\usepackage{graphicx}
\usepackage{booktabs}
\usepackage[utf8]{inputenc}
\usepackage{booktabs} 
\usepackage{multirow} 
\usepackage[table]{xcolor} 
\usepackage{adjustbox} 
\usepackage{amsmath} 
\usepackage{hyperref}
\usepackage{cleveref}

\begin{document}

\title{FTPFusion: Frequency-Aware Infrared and Visible Video Fusion with Temporal Perturbation}


\author{Xilai Li}
\email{20210300236@stu.fosu.edu.cn}
\affiliation{%
  \institution{Foshan University}
  \country{China}
}

\author{Chusheng Fang}
\email{2112555011@stu.fosu.edu.cn}
\affiliation{%
  \institution{Foshan University}
  \country{China}
}

\author{Xiaosong Li}
\authornote{Corresponding author.}
\email{lixiaosong@buaa.edu.cn}
\affiliation{%
  \institution{Foshan University}
  \country{China}
}

\renewcommand{\shortauthors}{Li et al.}
\renewcommand\footnotetextcopyrightpermission[1]{}
\settopmatter{printacmref=false} 
\begin{abstract}
Infrared and visible video fusion plays a critical role in intelligent surveillance and low-light monitoring. However, maintaining temporal stability while preserving spatial detail remains a fundamental challenge. Existing methods either focus on frame-wise enhancement with limited temporal modeling or rely on heavy spatio-temporal aggregation that often sacrifices high-frequency details. In this paper, we propose FTPFusion, a frequency-aware infrared and visible video fusion method based on temporal perturbation and sparse cross-modal interaction. Specifically, FTPFusion decomposes the feature representations into high-frequency and low-frequency components for collaborative modeling. The high-frequency branch performs sparse cross-modal spatio-temporal interaction to capture motion-related context and complementary details. The low-frequency branch introduces a temporal perturbation strategy to enhance robustness against complex video variations, such as flickering, jitter, and local misalignment. Furthermore, we design an offset-aware temporal consistency constraint to explicitly stabilize cross-frame representations under temporal disturbances. Extensive experiments on multiple public benchmarks demonstrate that FTPFusion consistently outperforms state-of-the-art methods across multiple metrics in both spatial fidelity and temporal consistency. The source code will be available at \href{https://github.com/ixilai/FTPFusion}{https://github.com/ixilai/FTPFusion}.
\end{abstract}




\maketitle

\section{Introduction}
Infrared and visible video fusion (IVVF) \cite{r137,r139,r142} aims to integrate complementary information from different sensors to generate fused videos with enhanced target saliency, semantic consistency, and temporal continuity. By providing more accurate and comprehensive scene semantics for downstream tasks, this technology plays a crucial role in intelligent surveillance and target perception in complex environments.

\begin{figure}[t]
  \centering
   \includegraphics[width=1\linewidth]{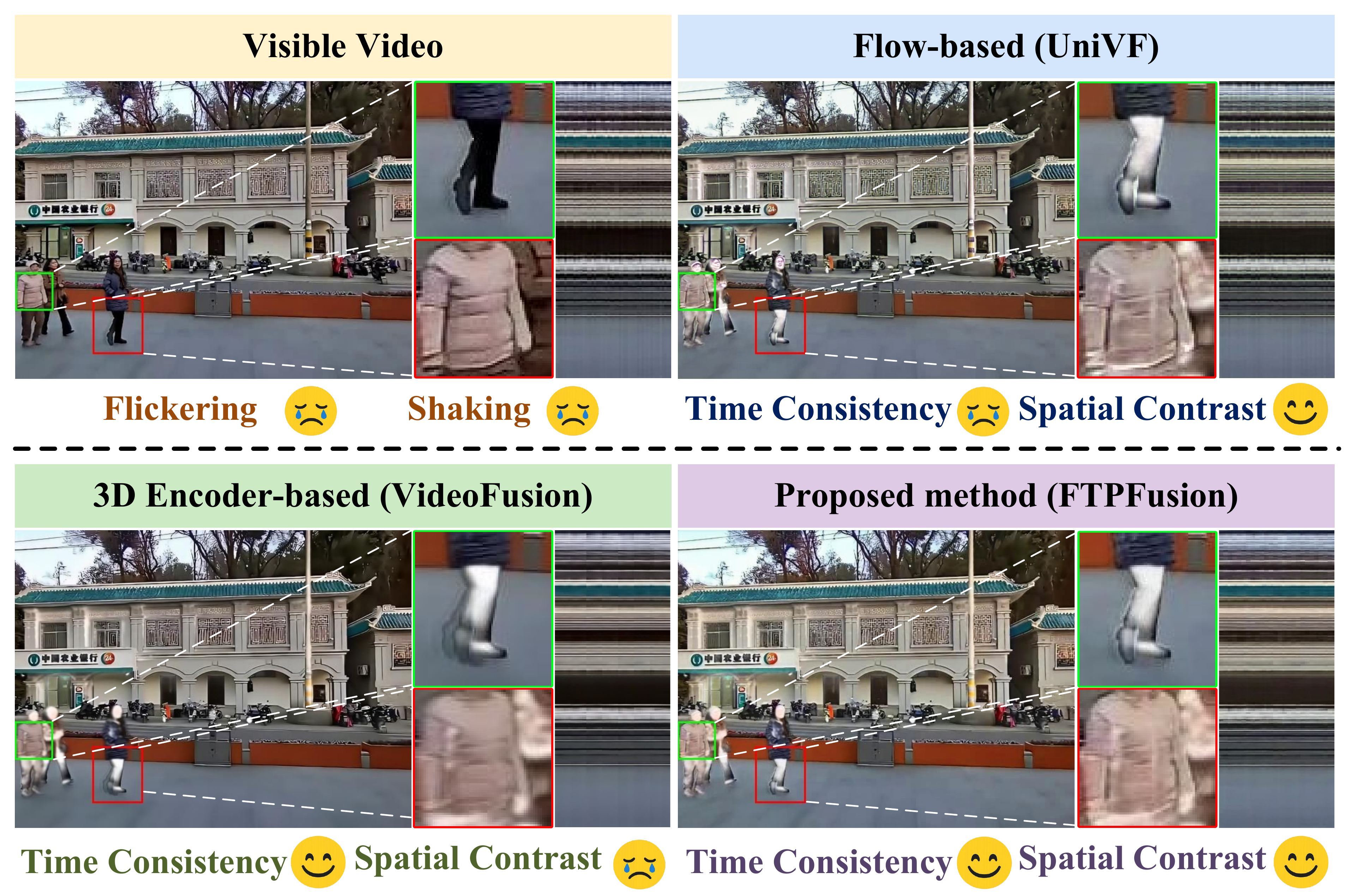}
   \caption{Comparison of the Limitations of Two Mainstream Video Fusion Methods.}
   \label{fig1}
\end{figure}

Compared to static image fusion, the primary challenge of IVVF lies not only in extending the fusion to video sequences but also in balancing spatial fidelity within individual frames and temporal consistency across frames \cite{r139,r134}. A frame-by-frame processing paradigm \cite{r107,r112,r152,r153,r88} tends to overlook the inherent temporal continuity of video, leading to artifacts such as flickering and motion discontinuity. In contrast, incorporating multi-frame information introduces new challenges, including cross-frame dynamic interference and difficulties in representation learning \cite{r157}. Thus, the core of video fusion is to fully exploit inter-frame dynamic clues while preserving the spatial details and contrast of the target frame. 

Existing video fusion methods can be broadly categorized into two paradigms.
The first paradigm \cite{r133,r137} extracts dynamic features from consecutive frames and utilizes optical flow to align adjacent features with the current frame; subsequently, multi-modal interaction and fusion are performed around this aligned reference. While these methods explicitly incorporate neighboring frame information, which represents an advancement over traditional image fusion, their underlying modeling remains a current-frame-centered reconstruction paradigm. In these methods, temporal cues are often introduced mainly through pre-fusion alignment or feature warping, rather than being deeply integrated throughout the entire feature encoding process. Consequently, the optimization tends to prioritize individual visual quality, which favors high contrast and sharp edges. As shown in \Cref{fig1}, some representative methods in this category may have difficulty fully exploiting temporal information under rapid changes or flickering, which can lead to temporal instability in complex scenes.
The second paradigm \cite{r139,r134} treats video clips as a spatio-temporal whole and jointly extracts features using 3D encoders and spatio-temporal attention. This paradigm aligns more closely with the inherent structure of video signals; consequently, it can directly utilize video context to enhance spatio-temporal continuity and scene representation in the fused features. Compared to the first category, this 3D spatio-temporal encoding framework is more effective at mitigating flickering and jitter issues commonly found in frame-by-frame fusion. However, this approach introduces a different challenge; specifically, to suppress the aforementioned artifacts, the model often continuously aggregates information from neighboring frames. As a result, the fused output tends toward temporal smoothing, which weakens the high-frequency textures of the current frame. Furthermore, as shown in \Cref{fig1}, cross-frame information is often not effectively filtered during large-scale displacements or rapid motion; therefore, misaligned information from adjacent frames is introduced into the current frame \cite{r156}, ultimately leading to ghosting artifacts and degraded visual quality.

In summary, although existing IVVF methods have made significant progress through either "alignment-based frame-by-frame fusion" or "joint spatio-temporal modeling," they still struggle to effectively balance spatial fidelity and temporal consistency.
The former maintains contrast and structures within individual frames but fails to fully capture spatio-temporal dependencies, while the latter utilizes inter-frame dynamic information more effectively but often sacrifices spatial details or introduces cross-frame artifacts under complex motion conditions.\textbf{ Therefore, establishing a better balance between temporal stability and spatial detail fidelity remains a critical and unresolved challenge in IVVF.}

To address this issue, we propose FTPFusion, a frequency-aware IVVF framework based on temporal perturbation and sparse cross-modal interaction. Unlike existing methods \cite{r139,r134} that model all spatio-temporal information within a unified representation space, we revisit the IVVF task from a frequency perspective. Specifically, we argue that temporal stability and spatial fidelity in video fusion should not be treated equally within the same feature subspace; instead, they should be modeled differentially according to the representational attributes of different frequency components. Based on this observation, we construct a dual-branch frequency-aware fusion framework that performs targeted spatio-temporal modeling and cross-modal interaction in low-frequency and high-frequency subspaces, respectively. In the high-frequency branch, we design a Sparse Cross-modal Attention Module (SCAM) to perform sparse spatio-temporal interactions only on high-frequency components, which selectively fuses cross-frame motion cues and complementary details. This approach effectively maintains the spatial contrast of the current frame while utilizing neighboring frame context. For the low-frequency branch, we design a Low-Frequency Temporal Perturbation Enhancement Module (LFPM) that applies perturbations to low-frequency sequences along the temporal dimension during training; combined with low-frequency spatio-temporal optimization, this enables the model to handle real-world degradations such as flickering, slight jitter, and local misalignment. Low-frequency components govern global energy and structural changes across frames. Modeling their stability directly promotes temporal continuity. Additionally, we introduce an Offset-aware Temporal Consistency constraint to enhance the perception of model of real-world challenges, such as jitter and misalignment.
The primary contributions of the proposed method can be summarized as follows:

\begin{itemize}
\item We propose a frequency-aware framework for infrared and visible video fusion that explicitly decouples spatial detail preservation and temporal stability in the frequency domain, providing a new perspective for balancing these two competing objectives.

\item We construct a dual-branch frequency fusion framework, where the low-frequency branch enhances temporal stability via temporal perturbation, and the high-frequency branch captures motion-related details and cross-modal complementary information through sparse spatio-temporal interaction.

\item We propose an offset-aware temporal consistency loss to explicitly model temporal disturbances, such as jitter and misalignment, providing effective supervision for learning more stable cross-frame representations during training.

\item Extensive experiments on multiple public datasets show that the proposed method achieves superior performance in both quantitative evaluation and visual quality, supporting its effectiveness in spatial fidelity and temporal consistency.
\end{itemize}

\section{Related Work}

\subsection{Multi-modality Image Fusion}

Multi-modal image fusion has achieved significant progress in recent years and has demonstrated high value in visual applications such as object perception and scene understanding \cite{r115,r117,r140,r12,r32}. Existing research primarily focuses on unified modeling paradigms centered on deep networks\cite{r24,r23,r114}; meanwhile, the focus of related studies has expanded from merely improving visual quality to consistency modeling for downstream tasks \cite{r116,r17,r143,r146} and robust fusion in complex scenes \cite{r129,r65,r144,r145,r154}. For example, to alleviate the conflict between cross-task generalization and source information fidelity in unified fusion frameworks, Li et al. \cite{r138} constructed a feature representation with stronger generalization capabilities by leveraging the semantic priors of DINOv3. Yi et al. \cite{r65} proposed a degradation-aware image fusion method that coordinates degradation processing and multi-modal interaction through text guidance. To build a unified image fusion framework suitable for adverse weather, Li et al. \cite{r91} further proposed AWFusion, which utilizes atmospheric scattering models to provide a unified description of different degradation conditions and achieves feature extraction oriented toward degradation separation. Although these methods have made significant progress in single-frame fusion, most of them lack the perception of temporal information; consequently, they struggle to directly meet the requirements of temporal continuity and dynamic consistency in video fusion tasks.

\begin{figure*}[t]
  \centering
   \includegraphics[width=1\linewidth]{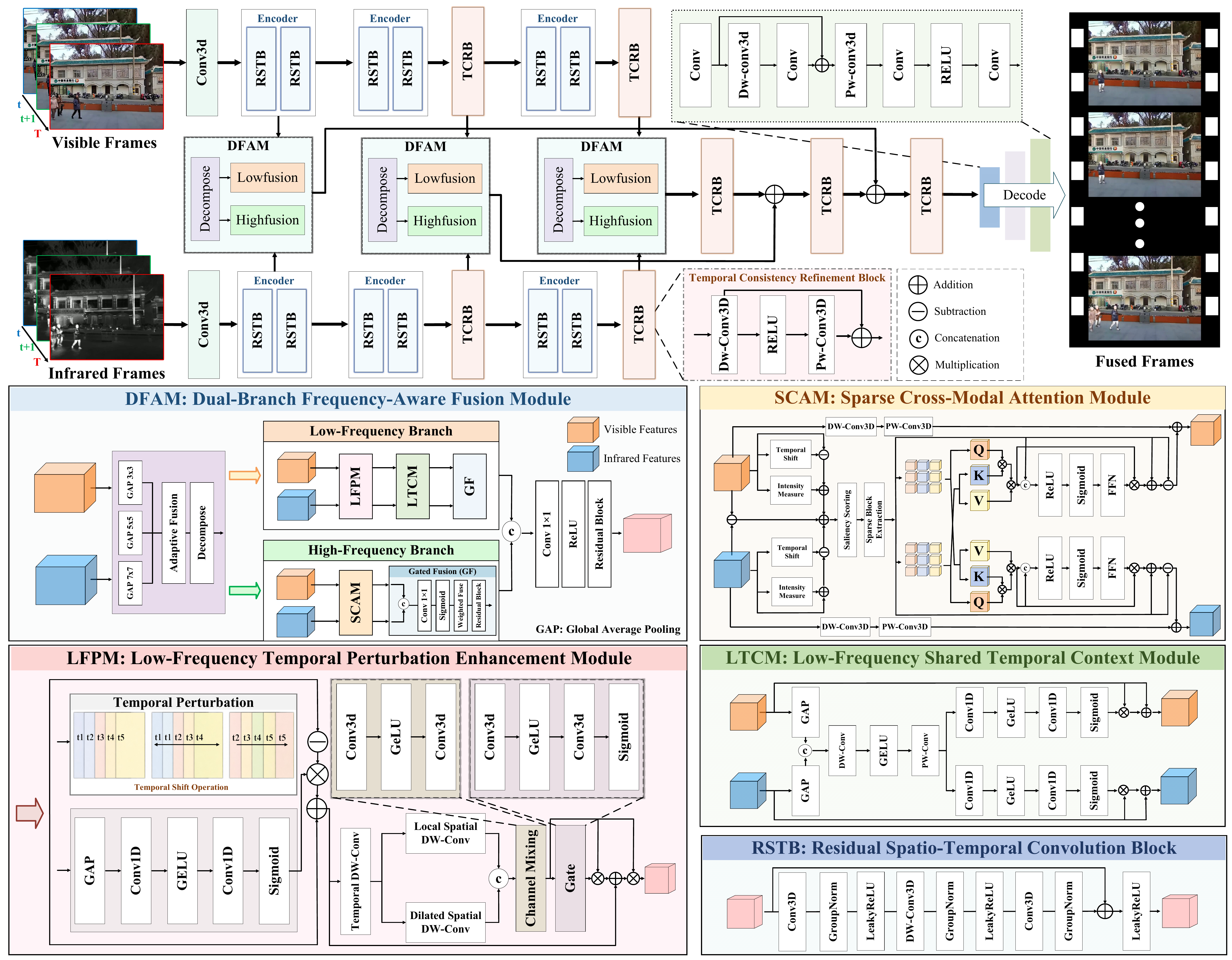}
   \caption{The overall framework of the proposed algorithm.}
   \label{fig2}
\end{figure*}

\subsection{ Multi-modality Video Fusion}

As more practical systems directly acquire long-sequence video data, multi-modal video fusion \cite{r136,r139,r134,r151} has received increasing attention in recent years. Compared to image fusion, video fusion requires not only integrating complementary information between different modalities but also modeling temporal dependencies within video sequences. Existing methods can be roughly categorized into two types. One category, represented by UniVF \cite{r133}, aligns neighboring frame information with the current frame through optical flow estimation and cross-frame feature warping; these methods subsequently combine temporal consistency losses to improve the continuity of fusion results. Although this category explicitly introduces multi-frame information, the fusion process still largely revolves around the aligned current frame. The other category employs a unified spatio-temporal modeling framework. For instance, VideoFusion \cite{r139} jointly models cross-modal complementary information and cross-frame dynamic information through a spatio-temporal collaborative network and a bidirectional temporal co-attention mechanism; similarly, TemCoCo \cite{r134} enhances temporal continuity in video fusion using 3D convolutions, temporal collaboration modules, and specialized temporal losses. Although these methods have advanced video fusion from frame-by-frame extension to explicit temporal modeling, how to maintain the spatial details of the current frame while fully utilizing inter-frame dynamic clues remains a problem worthy of further study.

\section{Methodology}
\subsection{Overview}

The overall framework of the proposed algorithm is shown in \Cref{fig2}. Given the registered infrared and visible video sequences, denoted as $\mathbf{X}^{ir}, \mathbf{X}^{vi} \in \mathbb{R}^{B \times T \times C \times H \times W}$, where $B$ represents the batch size, $T$ is the number of input frames, and $C$, $H$, and $W$ denote the number of channels, height, and width, respectively. The proposed FTPFusion uses an auto-encoder structure to jointly model the two modalities. Specifically, the network first extracts multi-scale features through modality-independent spatio-temporal encoders. Then, the Dual-Branch Frequency-Aware Fusion Module (DFAM) is introduced at each scale to perform frequency decomposition on infrared and visible features, followed by fusion in low-frequency and high-frequency subspaces. The low-frequency branch focuses on stable structures and slowly changing information shared across frames. It performs temporal enhancement and cross-modal context modeling using the Low-Frequency Temporal Perturbation Enhancement Module (LFPM) and the Low-Frequency Shared Temporal Context Module (LTCM). The high-frequency branch focuses on local changes related to textures, edges, and motion, employing the Sparse Cross-Modal Attention Module (SCAM) for targeted spatio-temporal interactions. After fusion, the network gradually restores spatial resolution through multi-scale decoding and temporal refinement to generate the final fused video.

\subsection{Dual-Branch Frequency-Aware Fusion Module}

Existing video fusion methods typically model all spatio-temporal information in a unified feature space, which limits the targeted fusion of different frequency components. To address this, we design a Dual-Branch Frequency-Aware Fusion Module (DFAM) to model and fuse infrared and visible features separately on the same scale.

For input infrared features $\mathbf{F}^{ir}$ and visible features $\mathbf{F}^{vi}$, DFAM first performs frequency decomposition on the 2D features of each frame, representing them as low-frequency (LF) and high-frequency (HF) components:
\begin{equation}
\left(\mathbf{L}^{ir}, \mathbf{H}^{ir}\right) = \mathcal{D}\left(\mathbf{F}^{ir}\right), \qquad \left(\mathbf{L}^{vi}, \mathbf{H}^{vi}\right) = \mathcal{D}\left(\mathbf{F}^{vi}\right),
\end{equation}
where $\mathcal{D}(\cdot)$ denotes the frequency decomposition operator. In implementation, LF components are constructed through multi-scale smoothing, where input features are filtered by average smoothing kernels of different scales and then aggregated using learnable weights. HF components are obtained by subtracting the LF components from the input features. After obtaining the LF and HF representations, DFAM processes them through the low-frequency and high-frequency branches. The low-frequency branch combines temporal perturbation enhancement and shared temporal context for fusion. The high-frequency branch models detailed information through sparse cross-modal spatio-temporal interaction. Finally, outputs from both branches are concatenated along the channel dimension, followed by convolutional mapping and feature reorganization to generate the fused features:
\begin{equation}
\mathbf{F}^{f} = \mathcal{M}\left(\left[\mathbf{F}^{low}, \lambda \mathbf{F}^{high}\right]\right),
\end{equation}
where $\mathbf{F}^{low}$ and $\mathbf{F}^{high}$ represent the fusion results of the low-frequency and high-frequency branches, respectively. $\lambda$ is a learnable high-frequency gain coefficient, and $\mathcal{M}(\cdot)$ denotes the final feature integration mapping.

\subsubsection{Low-Frequency Temporal Perturbation Enhancement Module}

The Low-Frequency Temporal Perturbation Enhancement Module (LFPM) is designed for temporal modeling in the low-frequency branch. It enhances the perception of slight jitter, flicker, and neighborhood misalignment in the low-frequency subspace. Given the input LF features $\mathbf{L} \in \mathbb{R}^{B \times T \times C \times H \times W}$, LFPM first performs average pooling along the spatial dimension to obtain the per-frame LF temporal representation $\mathbf{z}_t = \mathrm{Pool}(\mathbf{L}_t)$. A temporal activity map is then constructed based on the variation between adjacent frames. Specifically, the temporal activity response at frame $t$ is defined as:
\begin{equation}
\Delta_t =
\begin{cases}
\left| \mathbf{z}_{t+1} - \mathbf{z}_{t} \right|, & t=1, \\[2pt]
\frac{1}{2}\left( \left| \mathbf{z}_{t} - \mathbf{z}_{t-1} \right| + \left| \mathbf{z}_{t+1} - \mathbf{z}_{t} \right| \right), & 1 < t < T, \\[2pt]
\left| \mathbf{z}_{t} - \mathbf{z}_{t-1} \right|, & t=T.
\end{cases}
\end{equation}
Subsequently, a 1D temporal gating mapping generates the perturbation weights $\mathbf{G}=\sigma(\phi(\Delta))$ for each time step and channel, where $\phi(\cdot)$ is a temporal convolution and $\sigma(\cdot)$ is the Sigmoid activation.

LFPM introduces lightweight non-circular shifts along the temporal dimension during training to explicitly simulate temporal perturbations. Let $\mathcal{S}(\cdot)$ denote the temporal shift operator. The perturbed LF representation is expressed as:
\begin{equation}
\mathbf{L}^{p} = \mathbf{L} + \beta \mathbf{G} \odot \left( \mathcal{S}(\mathbf{L}) - \mathbf{L} \right),
\end{equation}
where $\beta$ is a learnable global perturbation strength and $\odot$ denotes element-wise multiplication. This process exposes the module to variations caused by temporal drift and misalignment during training, improving the robustness of LF representations.

After temporal perturbation, LFPM introduces a lightweight local 3D spatio-temporal enhancement branch to model local temporal dependencies. This branch applies depthwise separable convolutions along the temporal dimension and injects enhancement information via gated residuals. Let this mapping be $\mathcal{E}(\cdot)$. The enhanced LF features are represented as:
\begin{equation}
\tilde{\mathbf{L}} = \mathbf{L}^{p} + \gamma \odot \mathcal{E}\left(\mathbf{L}^{p}\right),
\end{equation}
where $\gamma$ is an injection coefficient modulated by temporal channel gating. To prevent the enhancement from causing excessive shifts in the LF base, the module constrains the consistency of the temporal mean at the output stage:
\begin{equation}
\hat{\mathbf{L}} = \tilde{\mathbf{L}} - \mathrm{Mean}_{t}\left(\tilde{\mathbf{L}}\right) + \mathrm{Mean}_{t}\left(\mathbf{L}\right).
\end{equation}
This design achieves joint optimization of temporal perturbation modeling and local spatio-temporal enhancement in the LF subspace.

\subsubsection{Low-Frequency Shared Temporal Context Module}

Apart from single-modality temporal enhancement, the Low-Frequency Shared Temporal Context Module (LTCM) models the shared temporal context between infrared and visible LF features. Given the processed LF features $\hat{\mathbf{L}}^{ir}$ and $\hat{\mathbf{L}}^{vi}$, LTCM extracts temporal tokens for both modalities via spatial average pooling. These tokens are concatenated along the channel dimension and mixed using a 1D depthwise separable temporal convolution to obtain the shared context:
\begin{equation}
\mathbf{z}^{s} = \psi\left(\left[\mathrm{Pool}\left(\hat{\mathbf{L}}^{ir}\right), \mathrm{Pool}\left(\hat{\mathbf{L}}^{vi}\right)\right]\right),
\end{equation}
where $\psi(\cdot)$ is the temporal mapping function.

Based on the shared temporal context, LTCM generates modulation gates for the infrared and visible branches and applies them to the corresponding LF features:
\begin{equation}
\bar{\mathbf{L}}^{ir} = \hat{\mathbf{L}}^{ir} \odot \left(1 + \sigma\left(\varphi_{ir}\left(\mathbf{z}^{s}\right)\right)\right), \qquad \bar{\mathbf{L}}^{vi} = \hat{\mathbf{L}}^{vi} \odot \left(1 + \sigma\left(\varphi_{vi}\left(\mathbf{z}^{s}\right)\right)\right),
\end{equation}
where $\varphi_{ir}(\cdot)$ and $\varphi_{vi}(\cdot)$ are mapping functions for each modality. Finally, the modulated representations are integrated via a gated fusion mapping:
\begin{equation}
\mathbf{F}^{low} = \mathrm{Fuse}_{low}\left(\bar{\mathbf{L}}^{ir}, \bar{\mathbf{L}}^{vi}\right).
\end{equation}
This process injects shared temporal context into LF modeling, ensuring that the fusion results maintain stable structures and cross-modal coordination.

\subsubsection{SCAM: Sparse Cross-Modal Attention Module}

The Sparse Cross-Modal Attention Module (SCAM) is employed in the high-frequency branch to perform cross-modal spatio-temporal interactions on discriminative regions while controlling computational costs. Given the HF features $\mathbf{H}^{ir}, \mathbf{H}^{vi} \in \mathbb{R}^{B \times T \times C \times H \times W}$, SCAM first constructs a joint importance score map $\mathbf{S}$ to measure the high-frequency saliency of each spatio-temporal position:
\begin{equation}
\mathbf{S} = \left| \mathbf{H}^{ir} - \mathbf{H}^{ir}_{prev} \right| + \left| \mathbf{H}^{vi} - \mathbf{H}^{vi}_{prev} \right| + \frac{1}{2}\left( \left| \mathbf{H}^{ir} \right| + \left| \mathbf{H}^{vi} \right| \right) + \left| \mathbf{H}^{ir} - \mathbf{H}^{vi} \right|,
\end{equation}
where $\mathbf{H}_{prev}$ denotes the features from the previous time step.

The module then partitions the score map into blocks and performs average aggregation to obtain importance scores for each spatial block. Given the total number of blocks $N_b$, only the top $K$ blocks are selected for interaction, where:
\begin{equation}
K = \max\left(1, \left\lfloor \rho N_b \right\rfloor \right),
\end{equation}
and $\rho$ is the sparse selection ratio. For the selected high-importance blocks, SCAM flattens them into token sequences and generates queries, keys, and values through linear mappings. Explicit spatio-temporal position encodings are introduced based on block centers to enhance awareness of spatial location and temporal order.

SCAM performs bidirectional cross-modal attention, where infrared tokens query visible tokens and vice versa. The outputs are expressed as:
\begin{equation}
\mathbf{A}^{ir \leftarrow vi} = \mathrm{Attn}\left(\mathbf{q}^{ir}, \mathbf{k}^{vi}, \mathbf{v}^{vi}\right), \qquad \mathbf{A}^{vi \leftarrow ir} = \mathrm{Attn}\left(\mathbf{q}^{vi}, \mathbf{k}^{ir}, \mathbf{v}^{ir}\right),
\end{equation}
where $\mathrm{Attn}(\mathbf{q}, \mathbf{k}, \mathbf{v}) = \mathrm{Softmax}\left(\mathbf{q}\mathbf{k}^{\top} / \sqrt{d}\right)\mathbf{v}$. After obtaining the attention results, the high-frequency representations of the selected blocks are updated via gated residuals and refined by feed-forward mappings. The updated tokens are then written back to their original spatial positions. 

In addition to sparse attention, SCAM maintains a local 3D convolutional branch for local spatio-temporal modeling of both modalities. The final HF fusion result is:
\begin{equation}
\mathbf{F}^{high} = \mathrm{Fuse}_{high}\left(\mathbf{H}^{ir}_{attn} + \mathcal{L}\left(\mathbf{H}^{ir}\right), \mathbf{H}^{vi}_{attn} + \mathcal{L}\left(\mathbf{H}^{vi}\right)\right),
\end{equation}
where $\mathcal{L}(\cdot)$ denotes the local 3D spatio-temporal modeling mapping. This design focuses on local dynamics and complementary information in the HF subspace while reducing computational redundancy.

\subsection{Loss Functions}

To simultaneously constrain the spatial information preservation and temporal consistency of the fused results, we use a composite objective function consisting of intensity loss, gradient loss, color loss \cite{r139,r65}, and Offset-aware Temporal Consistency loss:
\begin{equation}
\mathcal{L}_{total} = \lambda_{int}\mathcal{L}_{int} + \lambda_{grad}\mathcal{L}_{grad} + \lambda_{color}\mathcal{L}_{color} + \lambda_{tc}\mathcal{L}_{tc},
\end{equation}
where $\lambda_{int}$, $\lambda_{grad}$, $\lambda_{color}$, and $\lambda_{tc}$ denote the weight coefficients for each loss term.

\subsubsection{Offset-aware Temporal Consistency Loss}

Beyond image-level losses, we design an Offset-aware Temporal Consistency Loss $\mathcal{L}_{tc}$ to improve the capability of the model to handle slight jitter and local misalignment. Given the infrared input sequence $\mathbf{X}^{ir}$, visible input sequence $\mathbf{X}^{vi}$, and fused output sequence $\hat{\mathbf{Y}}$, we first map them into the grayscale space and apply low-pass filtering to obtain low-frequency representations, followed by frame-wise normalization. For a modality index $m \in \{ir, vi, f\}$, this is expressed as:
\begin{equation}
\mathbf{L}^{m}=\mathcal{G}(\mathbf{X}^{m}), \qquad \bar{\mathbf{L}}^{m}=\frac{\mathbf{L}^{m}-\mu(\mathbf{L}^{m})}{\sigma(\mathbf{L}^{m})},
\end{equation}
where $\mathbf{X}^{f}=\hat{\mathbf{Y}}$ and $\mathcal{G}(\cdot)$ denotes the low-pass filtering operation.

Let the center frame index be $c=\lfloor T/2 \rfloor$. For any neighboring frame $i \neq c$, we estimate the 2D translation offset relative to the center frame within a local search window. A reference offset is constructed using the matching confidence of both infrared and visible modalities:
\begin{equation}
\Delta_{m}^{i \rightarrow c}=\mathcal{M}\left(\bar{\mathbf{L}}_{i}^{m}, \bar{\mathbf{L}}_{c}^{m}\right), \qquad \Delta_{ref}^{i \rightarrow c}=\omega_{vi}^{i}\Delta_{vi}^{i \rightarrow c}+\omega_{ir}^{i}\Delta_{ir}^{i \rightarrow c},
\end{equation}
where $\mathcal{M}(\cdot,\cdot)$ is the offset estimation function and $\omega_{vi}^{i}+\omega_{ir}^{i}=1$.

Based on the reference offset, we align the neighboring low-frequency frames to the center frame and construct the fused sequence residual and the reference residual:
\begin{equation}
\begin{aligned}
\mathbf{R}_{f}^{i} &= \mathcal{W}\left(\mathbf{L}_{i}^{f}, \Delta_{ref}^{i \rightarrow c}\right) - \mathbf{L}_{c}^{f}, \\
\mathbf{R}_{ref}^{i} &= \omega_{vi}^{i} \left( \mathcal{W}\left(\mathbf{L}_{i}^{vi}, \Delta_{ref}^{i \rightarrow c}\right) - \mathbf{L}_{c}^{vi} \right) + \omega_{ir}^{i} \left( \mathcal{W}\left(\mathbf{L}_{i}^{ir}, \Delta_{ref}^{i \rightarrow c}\right) - \mathbf{L}_{c}^{ir} \right),
\end{aligned}
\end{equation}
where $\mathcal{W}(\cdot,\cdot)$ denotes the offset-based resampling operation.

The final $\mathcal{L}_{tc}$ consists of an offset supervision term, a low-frequency residual alignment term, and a residual gradient consistency term:
\begin{equation}
\begin{aligned}
\mathcal{L}_{tc} = \, & \lambda_{shift} \frac{1}{T-1} \sum_{i \neq c} \mathcal{L}_{\delta} \left(\Delta_{f}^{i \rightarrow c} - \Delta_{ref}^{i \rightarrow c}\right) \\
& + \lambda_{align} \frac{1}{T-1} \sum_{i \neq c} \rho \left(\mathbf{R}_{f}^{i} - \mathbf{R}_{ref}^{i}\right) \\
& + \lambda_{g} \frac{1}{T-1} \sum_{i \neq c} \rho \left(\nabla \mathbf{R}_{f}^{i} - \nabla \mathbf{R}_{ref}^{i}\right),
\end{aligned}
\end{equation}
where $\mathcal{L}_{\delta}(\cdot)$ denotes the Huber loss (also known as smooth $\ell_1$ loss), $\rho(\cdot)$ is the Charbonnier penalty function, $\nabla$ is the spatial gradient operator, and $\lambda_{shift}$, $\lambda_{align}$, and $\lambda_{g}$ are balancing coefficients.
Overall, $\mathcal{L}_{int}$, $\mathcal{L}_{grad}$, and $\mathcal{L}_{color}$ constrain the fused results from the perspective of spatial content preservation. $\mathcal{L}_{tc}$ provides supplementary supervision for low-frequency temporal alignment and dynamic consistency, helping the fused video achieve a better balance between spatial details and temporal continuity.

\begin{figure*}[t]
  \centering
   \includegraphics[width=1\linewidth]{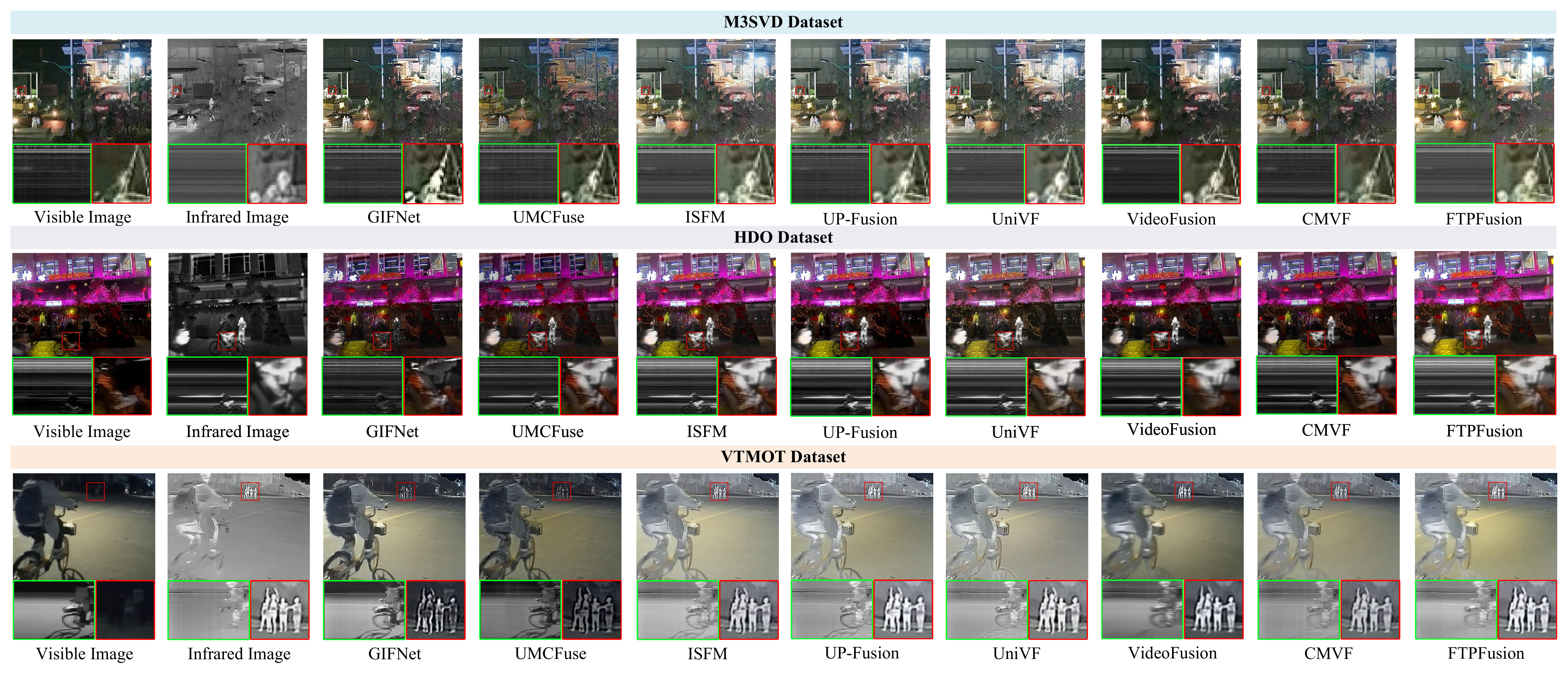}
   \caption{Qualitative comparison of all methods on the M3SVD, HDO and VTMOT datasets.}
   \label{fig3}
\end{figure*}

\section{Experiments}

\subsection{Training Sets}
\subsubsection{Comparison Methods and Datasets}

To validate the effectiveness of the proposed method, we select seven representative fusion approaches for comparison. These include four image fusion methods (GIFNet \cite{r114}, UMCFuse \cite{r129}, ISFM \cite{r147}, and UP-Fusion \cite{r140}) and three video fusion methods (UniVF \cite{r133}, VideoFusion \cite{r139}, and CMVF \cite{r135}). 

Experiments are conducted on three public video fusion benchmarks: M3SVD \cite{r139}, HDO \cite{r142}, and VTMOT \cite{r148}. For each dataset, five video sequences are randomly selected for testing, and the remaining ones are used for training. We evaluate the proposed method under two settings. In the main paper, we report results under a controlled perturbation setting, where temporal offsets, frame-wise flicker, and slight spatial perturbations are added to the test sequences to assess robustness to flicker, jitter, and local misalignment. We also evaluate the original standard benchmark setting without additional perturbations, and present those results in the appendix.

\subsubsection{Evaluation Metrics}

We employ eight metrics to evaluate the fusion results in terms of spatial information preservation and temporal quality. $Q_{MI}$, $Q_{TE}$ \cite{r69}, $EN$ \cite{r155}, and $SCD$ \cite{r149} are used to measure information content, texture details, and modal complementarity. $BiSwE$ and $MS2R$ \cite{r133} reflect temporal continuity and stability. Additionally, we introduce two new metrics to further assess modal mixing and temporal correlation: Modal Mixing Continuity Index ($MMCI$) and Temporal Correlation Preservation Error ($TCPE$).

\noindent \textbf{Video Quality Evaluation Metrics:}
Beyond standard objective metrics, we further introduce $MMCI$ and $TCPE$ to evaluate temporal quality. Specifically, $MMCI$ measures the stability of modal mixing between adjacent frames. Given the infrared frame $X_t^{ir}$, visible frame $X_t^{vi}$, and fused frame $\hat{Y}_t$, we estimate a local mixing coefficient $\alpha_t$ within a spatial neighborhood $\Omega$:
\begin{equation}
\alpha_t = \mathrm{clip} \left( \frac{\sum_{\Omega}(X_t^{ir}-X_t^{vi})(\hat{Y}_t-X_t^{vi})}{\sum_{\Omega}(X_t^{ir}-X_t^{vi})^2 + \varepsilon}, 0, 1 \right).
\end{equation}
To improve robustness, dynamic residuals are computed on Gaussian-smoothed frames. The metric is defined as
\begin{equation}
MMCI = s(J_{r} + \lambda_{\alpha}J_{\alpha}),
\end{equation}
where $J_{\alpha}$ denotes the temporal variation of $\alpha_t$ between adjacent frames, $J_r$ measures the consistency of the corresponding dynamic residuals, $\lambda_{\alpha}$ is a balancing factor, and $s$ is a scaling coefficient. A smaller $MMCI$ indicates more stable cross-frame modal mixing.

$TCPE$ evaluates how well the fused sequence preserves the temporal trends of the source modalities. Within a sliding window $W$, we compute the temporal correlation between the fused sequence and the local contrast trajectory of the infrared sequence, as well as that between the fused sequence and the gradient trajectory of the visible sequence. It is defined as
\begin{equation}
TCPE = \frac{1}{N_W} \sum_{n=1}^{N_W} E_{W_n},
\end{equation}
with window-wise error
\begin{equation}
E_W = \frac{1}{|\Omega|} \sum_{\Omega} (w d_R + (1-w) d_V),
\end{equation}
where $d_R$ and $d_V$ are the correlation errors for infrared contrast and visible gradient variations, respectively, and $w$ is an adaptive weight determined by their relative strengths. A smaller $TCPE$ indicates better preservation of temporal variations. Detailed definitions of $J_{\alpha}$, $J_r$, $d_R$, $d_V$, $w$, and the corresponding window settings are provided in the appendix.

\subsubsection{Implementation Details}

The proposed model is implemented in PyTorch and trained on four NVIDIA RTX 3090 GPUs.  During the first 70\% of the training epochs, the input patch size is $128 \times 128$ with a batch size of 8. In the remaining stage, the patch size increases to $256 \times 256$ and the batch size is adjusted to 2. We use the AdamW optimizer with an initial learning rate of $1 \times 10^{-4}$ and a weight decay of $1 \times 10^{-5}$. The total training process consists of 30 epochs, including a warm-up period of 6000 iterations. Additional implementation details and hyperparameter settings related to the proposed method are provided in the appendix.

\begin{figure}[t]
  \centering
   \includegraphics[width=1\linewidth]{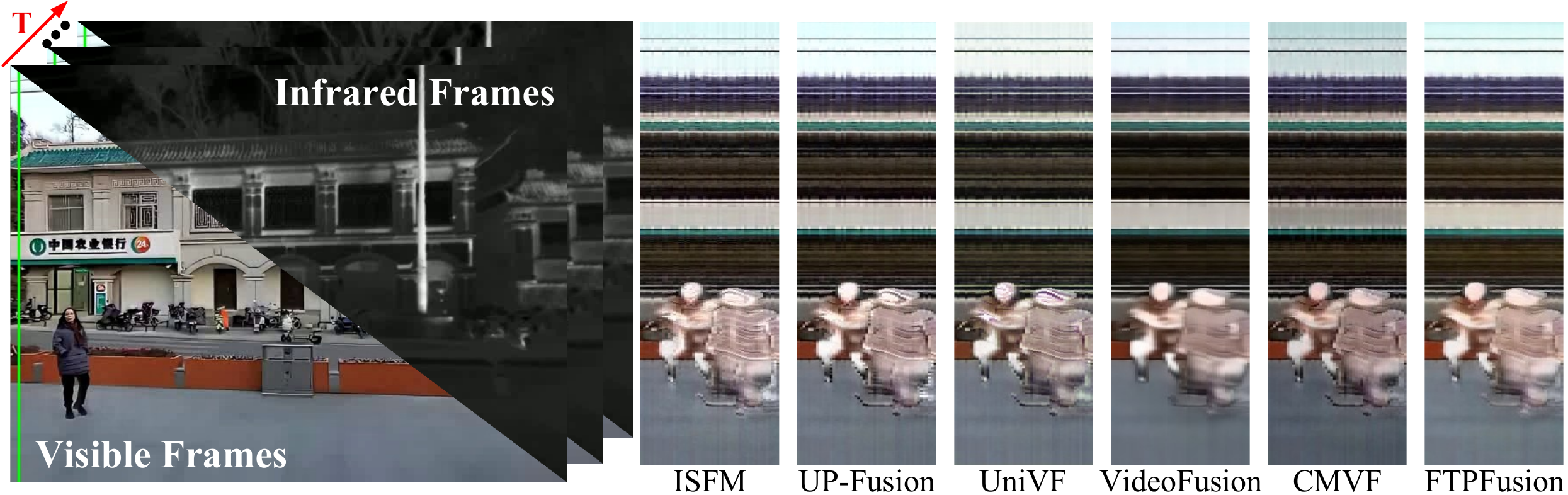}
   \caption{Visual comparison of the temporal stability of different methods.}
   \label{fig4}
\end{figure}

\definecolor{bestcolor}{HTML}{F4D9D7}
\definecolor{secondcolor}{HTML}{D9E1F2}
\begin{table*}[t]
\centering
\caption{Quantitative comparison on the M3SVD, HDO, and VTMOT datasets. The best results are highlighted in \colorbox{bestcolor}{bestcolor} and boldface, while the second-best results are highlighted in \colorbox{secondcolor}{secondcolor}.}
\label{tab1}
\begin{adjustbox}{max width=\textwidth}
\begin{tabular}{ccccccccccc}
\hline
\multicolumn{1}{c|}{Type}                    & \multicolumn{1}{c|}{Methods}     & \multicolumn{1}{c|}{Pub.}                & $Q_{MI}$↑                                     & $Q_{TE}$↑                                    & $EN$↑                                      & $SCD$↑                                     & $BiSwE$↓                                  & $MS2R$↓                                    & $MMCI$↓                                    & $TCPE$↓                                    \\ \hline
\multicolumn{11}{c}{\textbf{M3SVD Dataset}}                                                                                                                                                                                                                                                                                                                                                                                                                                \\ \hline
\multicolumn{1}{c|}{}                        & \multicolumn{1}{c|}{GIFNet}      & \multicolumn{1}{c|}{\textit{CVPR 25}}    & 0.3883                                  & 0.2955                                  & 6.9308                                  & 1.5724                                  & 8.5027                                  & \cellcolor[HTML]{D9E1F2}0.1758          & 0.2774                                  & 0.3809                                  \\
\multicolumn{1}{c|}{}                        & \multicolumn{1}{c|}{UMCFuse}     & \multicolumn{1}{c|}{\textit{TIP 25}}     & 0.4002                                  & 0.2972                                  & 6.8675                                  & 1.4233                                  & 7.6986                                  & 0.1873                                  & 0.2454                                  & 0.3539                                  \\
\multicolumn{1}{c|}{}                        & \multicolumn{1}{c|}{ISFM}        & \multicolumn{1}{c|}{\textit{TIP 26}}     & 0.5239                                  & 0.3804                                  & \cellcolor[HTML]{D9E1F2}7.3283          & 1.5966                                  & 8.2401                                  & 0.2032                                  & 0.2544                                  & 0.3362                                  \\
\multicolumn{1}{c|}{\multirow{-4}{*}{Image}} & \multicolumn{1}{c|}{UP-Fusion}   & \multicolumn{1}{c|}{\textit{AAAI 26}}    & 0.5774                                  & 0.3811                                  & 7.3126                                  & \cellcolor[HTML]{D9E1F2}1.6894          & 7.7609                                  & 0.2165                                  & 0.2608                                  & \cellcolor[HTML]{F4D9D7}\textbf{0.2749} \\ \hline
\multicolumn{1}{c|}{}                        & \multicolumn{1}{c|}{UniVF}       & \multicolumn{1}{c|}{\textit{NeurIPS 25}} & 0.5724                                  & \cellcolor[HTML]{D9E1F2}0.3928          & 7.3230                                  & 1.6802                                  & 6.4193                                  & 0.2152                                  & \cellcolor[HTML]{D9E1F2}0.2007          & 0.2764                                  \\
\multicolumn{1}{c|}{}                        & \multicolumn{1}{c|}{VideoFusion} & \multicolumn{1}{c|}{\textit{CVPR 26}}    & 0.5415                                  & 0.3658                                  & 7.3197                                  & 1.6274                                  & \cellcolor[HTML]{D9E1F2}5.9539          & 0.1869                                  & 0.2119                                  & 0.4086                                  \\
\multicolumn{1}{c|}{}                        & \multicolumn{1}{c|}{CMVF}        & \multicolumn{1}{c|}{\textit{INFFus 26}}  & \cellcolor[HTML]{D9E1F2}0.5781          & 0.3867                                  & 7.1590                                  & 1.6138                                  & 6.9036                                  & 0.2274                                  & 0.2434                                  & 0.2804                                  \\
\multicolumn{1}{c|}{\multirow{-4}{*}{Video}} & \multicolumn{1}{c|}{FTPFusion}    & \multicolumn{1}{c|}{\_}                  & \cellcolor[HTML]{F4D9D7}\textbf{0.5816} & \cellcolor[HTML]{F4D9D7}\textbf{0.4088} & \cellcolor[HTML]{F4D9D7}\textbf{7.3417} & \cellcolor[HTML]{F4D9D7}\textbf{1.6924} & \cellcolor[HTML]{F4D9D7}\textbf{5.4638} & \cellcolor[HTML]{F4D9D7}\textbf{0.1616} & \cellcolor[HTML]{F4D9D7}\textbf{0.1834} & \cellcolor[HTML]{D9E1F2}0.2755          \\ \hline
\multicolumn{11}{c}{\textbf{HDO Dataset}}                                                                                                                                                                                                                                                                                                                                                                                                                                  \\ \hline
\multicolumn{1}{c|}{}                        & \multicolumn{1}{c|}{GIFNet}      & \multicolumn{1}{c|}{\textit{CVPR 25}}    & 0.2921                                  & 0.2970                                  & 7.1004                                  & 1.5423                                  & 13.2702                                 & 1.1050                                  & 0.3383                                  & 0.4016                                  \\
\multicolumn{1}{c|}{}                        & \multicolumn{1}{c|}{UMCFuse}     & \multicolumn{1}{c|}{\textit{TIP 25}}     & 0.3382                                  & 0.2865                                  & 7.1135                                  & 1.2024                                  & 13.3657                                 & 1.0899                                  & 0.3334                                  & 0.3844                                  \\
\multicolumn{1}{c|}{}                        & \multicolumn{1}{c|}{ISFM}        & \multicolumn{1}{c|}{\textit{TIP 26}}     & 0.3941                                  & \cellcolor[HTML]{D9E1F2}0.3518          & 7.3474                                  & 1.5583                                  & 11.9769                                 & 1.0674                                  & 0.2248                                  & 0.3800                                  \\
\multicolumn{1}{c|}{\multirow{-4}{*}{Image}} & \multicolumn{1}{c|}{UP-Fusion}   & \multicolumn{1}{c|}{\textit{AAAI 26}}    & \cellcolor[HTML]{D9E1F2}0.4603          & 0.3411                                  & 7.3440                                  & 1.5520                                  & 11.3341                                 & 1.0698                                  & 0.2538                                  & 0.2898                                  \\ \hline
\multicolumn{1}{c|}{}                        & \multicolumn{1}{c|}{UniVF}       & \multicolumn{1}{c|}{\textit{NeurIPS 25}} & 0.4573                                  & 0.3444                                  & \cellcolor[HTML]{D9E1F2}7.3476          & \cellcolor[HTML]{F4D9D7}\textbf{1.6339} & 9.8693                                  & \cellcolor[HTML]{D9E1F2}1.0573          & 0.2219                                  & \cellcolor[HTML]{F4D9D7}\textbf{0.2752} \\
\multicolumn{1}{c|}{}                        & \multicolumn{1}{c|}{VideoFusion} & \multicolumn{1}{c|}{\textit{CVPR 26}}    & 0.4080                                  & 0.3348                                  & 7.2981                                  & 1.4372                                  & \cellcolor[HTML]{D9E1F2}9.8330          & 1.1555                                  & \cellcolor[HTML]{D9E1F2}0.2156          & 0.4250                                  \\
\multicolumn{1}{c|}{}                        & \multicolumn{1}{c|}{CMVF}        & \multicolumn{1}{c|}{\textit{INFFus 26}}  & 0.4049                                  & 0.3442                                  & 7.2795                                  & 1.4303                                  & 10.1597                                 & 1.0784                                  & 0.2755                                  & 0.3087                                  \\
\multicolumn{1}{c|}{\multirow{-4}{*}{Video}} & \multicolumn{1}{c|}{FTPFusion}    & \multicolumn{1}{c|}{\_}                  & \cellcolor[HTML]{F4D9D7}\textbf{0.4859} & \cellcolor[HTML]{F4D9D7}\textbf{0.3794} & \cellcolor[HTML]{F4D9D7}\textbf{7.3906} & \cellcolor[HTML]{D9E1F2}1.5845          & \cellcolor[HTML]{F4D9D7}\textbf{9.7316} & \cellcolor[HTML]{F4D9D7}\textbf{1.0448} & \cellcolor[HTML]{F4D9D7}\textbf{0.1873} & \cellcolor[HTML]{D9E1F2}0.2843          \\ \hline
\multicolumn{11}{c}{\textbf{VTMOT Dataset}}                                                                                                                                                                                                                                                                                                                                                                                                                                \\ \hline
\multicolumn{1}{c|}{}                        & \multicolumn{1}{c|}{GIFNet}      & \multicolumn{1}{c|}{\textit{CVPR 25}}    & 0.3448                                  & 0.4247                                  & 6.5272                                  & 1.5260                                  & 8.9334                                  & 0.8221                                  & 0.4266                                  & 0.3871                                  \\
\multicolumn{1}{c|}{}                        & \multicolumn{1}{c|}{UMCFuse}     & \multicolumn{1}{c|}{\textit{TIP 25}}     & 0.4359                                  & 0.4165                                  & 6.6452                                  & 1.3262                                  & 7.0099                                  & 0.8145                                  & 0.3266                                  & 0.3451                                  \\
\multicolumn{1}{c|}{}                        & \multicolumn{1}{c|}{ISFM}        & \multicolumn{1}{c|}{\textit{TIP 26}}     & 0.4331                                  & 0.4338                                  & \cellcolor[HTML]{D9E1F2}6.6700          & 1.4650                                  & 6.6871                                  & 0.8487                                  & 0.3234                                  & 0.3485                                  \\
\multicolumn{1}{c|}{\multirow{-4}{*}{Image}} & \multicolumn{1}{c|}{UP-Fusion}   & \multicolumn{1}{c|}{\textit{AAAI 26}}    & 0.5020                                  & 0.4255                                  & 6.6478                                  & 1.4766                                  & 6.7153                                  & 0.8159                                  & 0.3263                                  & \cellcolor[HTML]{D9E1F2}0.3016          \\ \hline
\multicolumn{1}{c|}{}                        & \multicolumn{1}{c|}{UniVF}       & \multicolumn{1}{c|}{\textit{NeurIPS 25}} & \cellcolor[HTML]{D9E1F2}0.5199          & 0.4267                                  & 6.5797                                  & \cellcolor[HTML]{D9E1F2}1.5293          & 6.2221                                  & \cellcolor[HTML]{D9E1F2}0.8123          & \cellcolor[HTML]{F4D9D7}\textbf{0.3113} & \cellcolor[HTML]{F4D9D7}\textbf{0.2995} \\
\multicolumn{1}{c|}{}                        & \multicolumn{1}{c|}{VideoFusion} & \multicolumn{1}{c|}{\textit{CVPR 26}}    & 0.4018                                  & \cellcolor[HTML]{F4D9D7}\textbf{0.4497} & 6.5432                                  & 1.3108                                  & 6.6803                                  & 1.2162                                  & 0.3584                                  & 0.4246                                  \\
\multicolumn{1}{c|}{}                        & \multicolumn{1}{c|}{CMVF}        & \multicolumn{1}{c|}{\textit{INFFus 26}}  & 0.4227                                  & 0.4420                                  & 6.5136                                  & 1.4994                                  & \cellcolor[HTML]{F4D9D7}\textbf{5.7174} & 0.8721                                  & 0.3179                                  & 0.3416                                  \\
\multicolumn{1}{c|}{\multirow{-4}{*}{Video}} & \multicolumn{1}{c|}{FTPFusion}    & \multicolumn{1}{c|}{\_}                  & \cellcolor[HTML]{F4D9D7}\textbf{0.5316} & \cellcolor[HTML]{D9E1F2}0.4472          & \cellcolor[HTML]{F4D9D7}\textbf{6.6893} & \cellcolor[HTML]{F4D9D7}\textbf{1.5366} & \cellcolor[HTML]{D9E1F2}6.0397          & \cellcolor[HTML]{F4D9D7}\textbf{0.8024} & \cellcolor[HTML]{D9E1F2}0.3164          & 0.3288                                  \\ \hline
\end{tabular}
\end{adjustbox}
\end{table*}

\subsection{Qualitative Comparative Experiment}
\Cref{fig3} illustrates the qualitative results of the proposed method compared with various image and video fusion approaches on the M3SVD, HDO, and VTMOT datasets. Red boxes highlight key local regions, while green boxes denote temporal slices. These slices are extracted from a fixed column position and stacked along the temporal dimension to visualize brightness variations and structural continuity across frames. Regarding spatial fidelity, image fusion methods like GIFNet and UMCFuse preserve some brightness and texture but exhibit blurriness and detail loss in the background. Video fusion methods such as VideoFusion and CMVF improve temporal continuity but often suffer from local over-smoothing or weakened textures. Temporal slices reveal that several image fusion methods still show noticeable stripe fluctuations and local jitter. The temporal slices of our method exhibit smoother and more natural structural transitions, which indicates better mitigation of inter-frame brightness jumps and local temporal instability. \Cref{fig4} further highlights these differences. Although visible images show significant flickering and jitter,  the proposed algorithm is still able to maintain the smoothness of the temporal transition. Overall, these visualizations demonstrate that while existing methods prioritize either spatial detail or temporal continuity, our approach achieves a better balance between the two.

\begin{table*}[t]
\centering
\caption{The ablation analysis of each component of the proposed method. The best results are highlighted in \colorbox{bestcolor}{bestcolor}.}
\label{tab2}
\begin{adjustbox}{max width=\textwidth}
\begin{tabular}{c|cccccccc}
\hline
Variants             & $Q_{MI}$↑                                     & $Q_{TE}$↑                                    & $EN$↑                                      & $SCD$↑                                     & $BiSwE$↓                                  & $MS2R$↓                                    & $MMCI$↓                                    & $TCPE$↓                                   \\ \hline
w/o DFAM            & 0.5324                                  & 0.3893                                  & 7.1067                                  & 1.6228                                  & 4.4204                                  & 0.1223                                  & 0.1807                                  & 0.3795                                  \\
High-frequency Only & 0.5487                                  & 0.4036                                  & 7.1089                                  & 1.7203                                  & 4.3024                                  & 0.1201                                  & 0.1790                                  & 0.3659                                  \\
Low-frequency Only  & 0.5974                                  & 0.3916                                  & 7.1064                                  & 1.6282                                  & 4.2825                                  & 0.1196                                  & 0.1765                                  & 0.3349                                  \\
w/o $\mathcal{L}_{tc}$             & 0.6125                                  & 0.4162                                  & 7.1142                                  & 1.7594                                  & 4.3709                                  & 0.1209                                  & 0.1783                                  & 0.3372                                  \\
Full Model          & \cellcolor[HTML]{F4D9D7}\textbf{0.6374} & \cellcolor[HTML]{F4D9D7}\textbf{0.4286} & \cellcolor[HTML]{F4D9D7}\textbf{7.1333} & \cellcolor[HTML]{F4D9D7}\textbf{1.7811} & \cellcolor[HTML]{F4D9D7}\textbf{4.2337} & \cellcolor[HTML]{F4D9D7}\textbf{0.1182} & \cellcolor[HTML]{F4D9D7}\textbf{0.1697} & \cellcolor[HTML]{F4D9D7}\textbf{0.3121} \\ \hline
\end{tabular}
\end{adjustbox}
\end{table*}

\begin{figure}[t]
  \centering
   \includegraphics[width=1\linewidth]{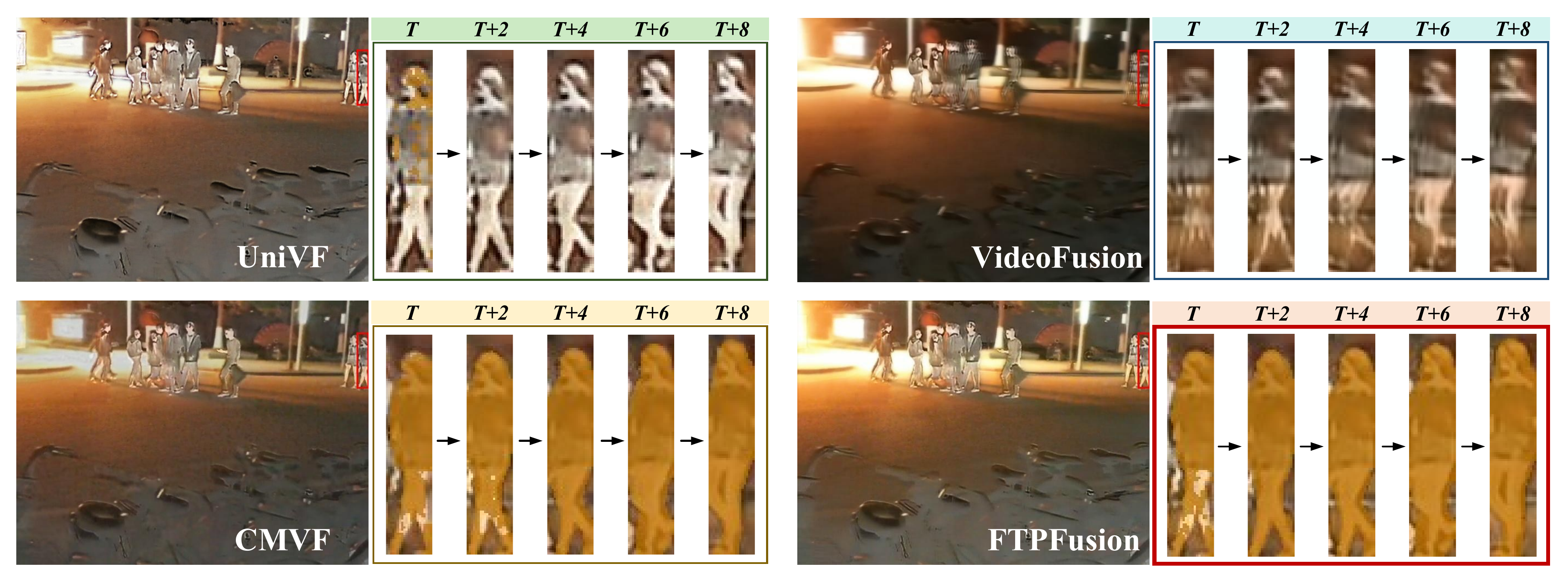}
   \caption{Comparison of segmentation results of different fusion methods in the video
segmentation task.}
   \label{fig5}
\end{figure}

\subsection{Quantitative comparative experiment}
\Cref{tab1} presents the quantitative comparison of different image and video fusion methods on the M3SVD, HDO, and VTMOT datasets. As shown in the table, our method achieves stable and competitive performance across all three datasets. In particular, it performs favorably on several spatial quality metrics, including $Q_{MI}$, $Q_{TE}$, $EN$, and $SCD$, suggesting its effectiveness in preserving salient infrared information and visible texture details. In terms of temporal quality, the proposed method also obtains competitive results on $BiSwE$, $MS2R$, $MMCI$, and $TCPE$, indicating its potential to alleviate flickering and cross-frame discontinuities in video fusion. The advantages of the proposed framework are more evident on the M3SVD and HDO datasets, while it still maintains high spatial quality on the more dynamic VTMOT dataset. Overall, these results suggest that our method provides a reasonable balance between cross-modal information extraction and temporal modeling, with promising robustness across diverse scenarios.

\subsection{Downstream Task Experiments}
Video segmentation results further validate the effectiveness of the proposed method in preserving salient semantic information and maintaining smooth transitions between frames. With SAMWISE \cite{r150} as the segmenter, \Cref{fig5} visualizes the segmentation of the rightmost pedestrian across consecutive time steps. While UniVF identifies partial target regions initially, its segmentation stability decreases over time. VideoFusion suffers from blurred contours and missing regions due to noticeable jitter and flickering, and CMVF preserves pedestrian semantics but still yields blurry regions and incomplete boundaries in early frames. In contrast, our method maintains more complete object regions and clearer boundaries across consecutive frames, providing more stable fused representations for downstream video segmentation.

\subsection{Ablation Study}

To evaluate the contribution of each component, we conduct ablation experiments on two sequences from M3SVD. We test the dual-branch frequency-aware fusion module, the low-frequency branch, the high-frequency branch, and the Offset-aware Temporal Consistency loss. The results are summarized in \Cref{tab2}.

\noindent\textbf{w/o DFAM.} 
We replace DFAM at all three scales with fusion blocks in a unified feature space, while keeping the other settings unchanged. This variant yields the largest performance drop, indicating that unified spatio-temporal modeling is insufficient to balance spatial detail preservation and temporal stability. In contrast, explicit frequency decomposition and dual-branch modeling are crucial to the full framework.

\noindent\textbf{Low-frequency Only.} 
In this variant, we retain the frequency decomposition and the full low-frequency branch while removing sparse cross-modal attention and high-frequency fusion. This variant performs best on temporal metrics, indicating that the low-frequency branch plays a key role in modeling stable structures and temporal continuity. However, its lower spatial metrics indicate that low-frequency modeling alone is insufficient for recovering textures and complementary details.

\noindent\textbf{High-frequency Only.} 
We keep the frequency decomposition and the full high-frequency branch while removing the low-frequency branch. Compared with Low-frequency Only, this variant performs better on spatial detail metrics but worse on temporal metrics, indicating that the high-frequency branch mainly benefits texture, edge, and local structure modeling, while low-frequency modeling remains essential for cross-frame stability.

\noindent\textbf{w/o $L_{tc}$.} 
To verify the effectiveness of the proposed $L_{tc}$, we remove it and train the model with only intensity, gradient, and color losses, while keeping all other settings unchanged. Removing $L_{tc}$ causes only minor changes in spatial metrics but consistently degrades temporal metrics, indicating that structural temporal modeling alone is insufficient for cross-frame continuity. The proposed $L_{tc}$ provides effective low-frequency temporal supervision and further improves temporal stability.

\subsection{Computational Efficiency Analysis}
Regarding computational efficiency, our method is substantially lighter than existing video fusion models. For \(480\mathrm{P}\) video, it requires only \(1.1\mathrm{M}\) parameters and \(132.149\mathrm{G}\) FLOPs, whereas VideoFusion uses \(6.7\mathrm{M}\) parameters and \(1874\mathrm{G}\) FLOPs, and UniVF further increases to \(9.2\mathrm{M}\) parameters and \(2164.07\mathrm{G}\) FLOPs. In other words, our model reduces the parameter count to about one-sixth of VideoFusion and nearly one-eighth of UniVF. 

\section{Conclusion}
FTPFusion is introduced to address the challenge of balancing spatial detail preservation and temporal consistency in infrared and visible video fusion. The method models video fusion from a frequency perspective, where low-frequency stable information and high-frequency details are processed separately within a unified framework. It also incorporates offset-aware temporal consistency constraints to enhance temporal robustness in complex dynamic scenarios. Experiments on multiple public datasets demonstrate that FTPFusion improves both the visual quality and temporal continuity of fused videos. Future work will focus on improving robustness in more challenging dynamic conditions.

\section*{Acknowledgement}
This research was supported by the National Natural Science Foundation of China (No. 62201149), the Natural Science Foundation of Guangdong Province (No. 2024A1515011880), and the Basic and Applied Basic Research of Guangdong Province (No. 2023A1515140077).

\bibliographystyle{ACM-Reference-Format}
\bibliography{sample-base}

\clearpage
\appendix

\section{Additional Implementation Details}
\label{app:details}

\subsection{Frequency decomposition}
In Eq.~(1), the decomposition operator $D(\cdot)$ is implemented by three fixed average-smoothing operators with kernel sizes
\begin{equation}
\mathcal{K}=\{3,5,7\}.
\end{equation}
For each scale $k\in\mathcal{K}$, reflection padding is applied before average pooling with stride $1$:
\begin{equation}
L^{(k)}=\mathrm{AvgPool}_{k}\!\left(\mathrm{Pad}_{k}(F)\right).
\end{equation}
Let $\{a_k\}$ denote the learnable scale logits. The normalized fusion weights are
\begin{equation}
\alpha_k=\frac{\exp(a_k)}{\sum_{j\in\mathcal{K}}\exp(a_j)}.
\end{equation}
The low-frequency component is computed as
\begin{equation}
L=\sum_{k\in\mathcal{K}}\alpha_k L^{(k)},
\end{equation}
and the high-frequency component is obtained by residual subtraction:
\begin{equation}
H=F-L.
\end{equation}
Hence, the smoothing kernels are fixed, while only the cross-scale aggregation weights are learned.

\subsection{Low-frequency branch}
The low-frequency perturbation module uses grouped temporal perturbation with $g=4$, maximum shift $s_{\max}=1$, perturbation probability $0.7$, and hidden ratio $0.25$. Symmetric non-circular temporal shifts are adopted during training, and temporal-mean preservation is enabled. The local spatio-temporal enhancement branch uses $3\times3\times3$ kernels together with a dilated spatial branch. These settings follow the implementation used in all experiments.

\subsection{SCAM}
In SCAM, the high-frequency features are partitioned into non-overlapping spatial blocks of size
\begin{equation}
p=8,
\end{equation}
with stride also equal to $8$. Therefore, blocks do not overlap spatially. At the three fusion stages, the sparse ratios are
\begin{equation}
\rho=\{0.10,0.20,0.30\},
\end{equation}
and the numbers of attention heads are
\begin{equation}
N_h=\{2,2,2\}.
\end{equation}
The attention expansion ratio is set to $0.75$ for all stages. Given the total number of blocks $N_b$, the number of selected blocks is
\begin{equation}
K=\max(1,\lfloor \rho N_b\rfloor).
\end{equation}
For each selected block, the tokens are flattened over the whole temporal clip, yielding a token length of $T\cdot p^2$. The attention dimension is
\begin{equation}
d=\mathrm{make\_divisible}\!\left(\max(\lfloor 0.75C\rfloor,N_h),N_h\right).
\end{equation}



\subsection{Operational Details of the Offset-aware Temporal Consistency Loss}
\label{app:tc_loss}

Let $V_f$, $V_{vi}$, and $V_{ir}$ denote the fused, visible, and infrared video clips, respectively. We impose temporal consistency only on the low-frequency components. Specifically, each frame is first converted to grayscale when $C=3$,
\begin{equation}
Y = 0.299R + 0.587G + 0.114B,
\end{equation}
and then filtered by a Gaussian kernel ($9\times9$, $\sigma=2.0$):
\begin{equation}
L_m = G_{\sigma}(V_m), \qquad m\in\{f,vi,ir\}.
\end{equation}
To reduce the influence of illumination variation, the low-frequency frames are normalized independently:
\begin{equation}
\bar{L}_m = \frac{L_m-\mu(L_m)}{\sigma(L_m)}.
\end{equation}

Let $c=\lfloor T/2\rfloor$ denote the center frame. For each neighboring frame $i\neq c$, we estimate the neighbor-to-center translation within a local search window $\mathcal{U}=[-2,2]^2$. For a candidate shift $(u,v)$, the matching cost is defined as
\begin{equation}
\mathcal{C}_{i\rightarrow c}^{m}(u,v)
=
\frac{1}{|\Omega|}
\sum_{\Omega}
\sqrt{\left(S_{u,v}(\bar{L}_i^{m})-\bar{L}_c^{m}\right)^2+\epsilon^2},
\end{equation}
where $m\in\{f,vi,ir\}$, $\epsilon=10^{-3}$, and $S_{u,v}(\cdot)$ denotes integer shift with replicate padding. Instead of directly selecting the best shift, we compute a soft probability over all candidates:
\begin{equation}
P_{i\rightarrow c}^{m}(u,v)
=
\frac{\exp\!\left(-\mathcal{C}_{i\rightarrow c}^{m}(u,v)/\tau\right)}
{\sum_{(u',v')\in\mathcal{U}}
\exp\!\left(-\mathcal{C}_{i\rightarrow c}^{m}(u',v')/\tau\right)},
\end{equation}
where $\tau=0.15$. The final sub-pixel shift is obtained by expectation:
\begin{equation}
\Delta_{i\rightarrow c}^{m}
=
\sum_{(u,v)\in\mathcal{U}} P_{i\rightarrow c}^{m}(u,v)\,[u,v].
\end{equation}

The reliability of each modality is measured from the entropy of the shift distribution, and the visible/infrared reference shift is computed by confidence-weighted fusion:
\begin{equation}
\omega_i^{vi},\omega_i^{ir}
=
\mathrm{Softmax}\!\left(
\gamma[\mathrm{conf}_i^{vi},\mathrm{conf}_i^{ir}]
\right),
\qquad \gamma=8.0,
\end{equation}
\begin{equation}
\Delta_{i\rightarrow c}^{ref}
=
\omega_i^{vi}\Delta_{i\rightarrow c}^{vi}
+
\omega_i^{ir}\Delta_{i\rightarrow c}^{ir}.
\end{equation}

Based on the reference shift, we supervise the fused-sequence shift by
\begin{equation}
L_{\mathrm{shift}}
=
\frac{1}{T-1}\sum_{i\neq c}
\mathrm{SmoothL1}\!\left(\Delta_{i\rightarrow c}^{f},\Delta_{i\rightarrow c}^{ref}\right).
\end{equation}

We then align neighboring low-frequency frames using bilinear warping with border padding:
\begin{equation}
W(X,\Delta)=\mathrm{GridSample}(X,\Delta).
\end{equation}
The aligned residuals are defined as
\begin{equation}
R_i^{f}=W(L_i^{f},\Delta_{i\rightarrow c}^{ref})-L_c^{f},
\end{equation}
\begin{equation}
R_i^{ref}
=
\omega_i^{vi}\!\left(W(L_i^{vi},\Delta_{i\rightarrow c}^{ref})-L_c^{vi}\right)
+
\omega_i^{ir}\!\left(W(L_i^{ir},\Delta_{i\rightarrow c}^{ref})-L_c^{ir}\right).
\end{equation}
The alignment term is
\begin{equation}
L_{\mathrm{align}}
=
\frac{1}{T-1}\sum_{i\neq c}
\frac{1}{|\Omega|}
\sum_{\Omega}
\sqrt{(R_i^f-R_i^{ref})^2+\epsilon^2}.
\end{equation}

To further improve sensitivity to slight local misalignment, we additionally constrain the gradients of the aligned residuals:
\begin{equation}
L_{\mathrm{grad}}
=
\frac{1}{2(T-1)}
\sum_{i\neq c}
\left(
\|\nabla_x R_i^f-\nabla_x R_i^{ref}\|_{\mathrm{char}}
+
\|\nabla_y R_i^f-\nabla_y R_i^{ref}\|_{\mathrm{char}}
\right).
\end{equation}

The final Offset-aware Temporal Consistency loss is defined as
\begin{equation}
L_{tc}
=
2.0\,L_{\mathrm{shift}}
+
1.0\,L_{\mathrm{align}}
+
0.3\,L_{\mathrm{grad}}.
\end{equation}

\subsection{Details of MMCI and TCPE}
\label{app:metrics}

\paragraph{MMCI}
The Modal Mixing Continuity Index (MMCI) measures the temporal stability of modal mixing. For frame $t$, let $X_t^{ir}$, $X_t^{vi}$, and $\hat{Y}_t$ denote the infrared, visible, and fused frames. A local pixel-wise mixing coefficient is estimated within neighborhood $\Omega$ by
\begin{equation}
\alpha_t
=
\mathrm{clip}\!\left(
\frac{\sum_{\Omega}(X_t^{ir}-X_t^{vi})(\hat{Y}_t-X_t^{vi})}
{\sum_{\Omega}(X_t^{ir}-X_t^{vi})^2+\varepsilon},
0,1
\right).
\end{equation}
After Gaussian smoothing, the continuity and residual terms are
\begin{equation}
J_{\alpha}
=
\frac{1}{T-1}\sum_{t=2}^{T}\|\alpha_t-\alpha_{t-1}\|_1,
\end{equation}
\begin{equation}
J_r
=
\frac{1}{T-1}\sum_{t=2}^{T}
\left\|
\Delta \tilde{Y}_t
-
\bar{\alpha}_t \Delta \tilde{X}_t^{ir}
-
(1-\bar{\alpha}_t)\Delta \tilde{X}_t^{vi}
\right\|_1,
\end{equation}
where $\bar{\alpha}_t=\frac{1}{2}(\alpha_t+\alpha_{t-1})$. The final metric is
\begin{equation}
\mathrm{MMCI}
=
s\left(J_r+\lambda_{\alpha}J_{\alpha}\right).
\end{equation}
A smaller MMCI indicates more stable cross-frame modal mixing.

\paragraph{TCPE}
The Temporal Correlation Preservation Error (TCPE) evaluates whether the fused video preserves the temporal variation trends of the source sequences. For a temporal window $\mathcal{W}$, the local contrast and gradient descriptors are
\begin{equation}
C(X)=|X-\mathrm{Mean}_{\Omega}(X)|,\qquad
G(X)=\frac{|\partial_x X|+|\partial_y X|}{8}.
\end{equation}
Within each window, Pearson correlations are computed between infrared and fused contrast trajectories, and between visible and fused gradient trajectories:
\begin{equation}
\rho_R=\mathrm{Corr}\!\left(\{C(X_t^{ir})\}_{t\in\mathcal{W}},\{C(\hat{Y}_t)\}_{t\in\mathcal{W}}\right),
\end{equation}
\begin{equation}
\rho_V=\mathrm{Corr}\!\left(\{G(X_t^{vi})\}_{t\in\mathcal{W}},\{G(\hat{Y}_t)\}_{t\in\mathcal{W}}\right).
\end{equation}
The corresponding distortions are
\begin{equation}
d_R=\frac{1-\rho_R}{2}, \qquad
d_V=\frac{1-\rho_V}{2}.
\end{equation}
An adaptive modality weight is defined as
\begin{equation}
w
=
\frac{\mathrm{Mean}_{t\in\mathcal{W}}\, C(X_t^{ir})}
{\mathrm{Mean}_{t\in\mathcal{W}}\, C(X_t^{ir})+\mathrm{Mean}_{t\in\mathcal{W}}\, G(X_t^{vi})+\varepsilon}.
\end{equation}
The window-wise error is
\begin{equation}
E_{\mathcal{W}}
=
\frac{1}{|\Omega|}
\sum_{\Omega}
\left(
w\,d_R+(1-w)\,d_V
\right),
\end{equation}
and the final TCPE is
\begin{equation}
\mathrm{TCPE}
=
\frac{1}{N_W}\sum_{n=1}^{N_W}E_{\mathcal{W}_n}.
\end{equation}
A smaller TCPE indicates better preservation of temporal source correlations.

\subsection{Effectiveness Validation of the Proposed Temporal Metrics}
\label{app:metric_effectiveness}

To verify the effectiveness of the proposed temporal metrics and further validate the motivation behind our metric design, we construct a stress benchmark with six degradation levels under three corruption families, including \emph{modality drift}, \emph{temporal shuffle}, and \emph{mixed hard}. These settings progressively perturb the fused videos from different perspectives and are used to compare two existing temporal metrics, \emph{BiSWE} and \emph{MS2R}, with the proposed \emph{MMCI} and \emph{TCPE}. We report multiple criteria, including global rank correlation with severity, linear correlation, monotonicity, pairwise ordering accuracy, adjacent-level separation, and average rank.

\begin{table*}[t]
\centering
\caption{Effectiveness validation of temporal metrics under the proposed stress benchmark. Best results in each column are in bold.}
\label{tab:stress_metric_summary}
\resizebox{\textwidth}{!}{
\begin{tabular}{llccccccc}
\toprule
Mode & Metric & Global Spearman $\uparrow$ & Global Pearson $\uparrow$ & Mean Seq. Spearman $\uparrow$ & Monotonic Rate $\uparrow$ & Pairwise Acc. $\uparrow$ & Adjacent Sep. $\uparrow$ & Avg. Rank $\downarrow$ \\
\midrule
\multirow{4}{*}{Mixed Hard}
& BiSWE & 0.8848 & \textbf{0.8798} & \textbf{0.8720} & \textbf{0.8889} & \textbf{0.9926} & 2.9125 & \textbf{1.625} \\
& MMCI  & \textbf{0.9363} & 0.8626 & 0.8667 & 0.7778 & 0.9704 & \textbf{3.2218} & \textbf{1.625} \\
& TCPE  & 0.4450 & 0.4148 & 0.5840 & 0.6111 & 0.8519 & 0.7433 & 2.750 \\
& MS2R  & -0.0261 & -0.0753 & 0.2587 & 0.0000 & 0.5259 & 0.0081 & 4.000 \\
\midrule
\multirow{4}{*}{Modality Drift}
& MMCI  & \textbf{0.9794} & \textbf{0.9588} & \textbf{0.9573} & 0.8889 & 0.9926 & \textbf{3.3475} & \textbf{1.625} \\
& TCPE  & 0.9586 & 0.9560 & 0.9520 & \textbf{1.0000} & \textbf{1.0000} & 2.1464 & \textbf{1.625} \\
& BiSWE & 0.9683 & 0.9399 & \textbf{0.9573} & \textbf{1.0000} & \textbf{1.0000} & 2.5543 & 2.750 \\
& MS2R  & 0.8117 & 0.6456 & 0.8187 & 0.4444 & 0.8889 & 0.8817 & 4.000 \\
\midrule
\multirow{4}{*}{Temporal Shuffle}
& TCPE  & \textbf{0.9474} & \textbf{0.9413} & \textbf{0.9493} & \textbf{1.0000} & \textbf{1.0000} & \textbf{0.4395} & \textbf{1.125} \\
& MMCI  & 0.6893 & 0.6230 & 0.7160 & 0.7778 & 0.9407 & 0.1659 & 1.500 \\
& MS2R  & 0.5111 & 0.3753 & 0.5773 & 0.5556 & 0.8741 & 0.0890 & 2.625 \\
& BiSWE & 0.1834 & 0.2910 & 0.0560 & 0.1111 & 0.6222 & 0.0401 & 3.625 \\
\bottomrule
\end{tabular}
}
\end{table*}

As shown in Table~\ref{tab:stress_metric_summary}, the proposed metrics exhibit clear task-dependent advantages, which directly support the design motivation of MMCI and TCPE. Under \emph{modality drift}, MMCI achieves the best overall performance, including the highest global Spearman correlation (\textbf{0.9794}), the highest global Pearson correlation (\textbf{0.9588}), the highest mean sequence-wise Spearman (\textbf{0.9573}), and the largest adjacent-level separation (\textbf{3.3475}). This indicates that MMCI is highly sensitive to progressively increasing cross-modal temporal inconsistency. Such behavior is well aligned with its design goal, since MMCI explicitly measures the continuity of modal mixing across frames. Once the temporal correspondence between infrared and visible information is disturbed, MMCI responds more strongly than existing metrics.

Under \emph{temporal shuffle}, TCPE shows the clearest advantage. It achieves the best results in all major criteria, including global Spearman (\textbf{0.9474}), global Pearson (\textbf{0.9413}), mean sequence-wise Spearman (\textbf{0.9493}), monotonic rate (\textbf{1.0000}), pairwise ordering accuracy (\textbf{1.0000}), adjacent separation (\textbf{0.4395}), and average rank (\textbf{1.125}). This is also consistent with the design of TCPE, which evaluates whether the fused sequence preserves the temporal variation trends of the source modalities. Therefore, when temporal order is disrupted, TCPE provides the most reliable response among all compared metrics.

Under \emph{mixed hard} corruption, BiSWE and MMCI obtain the same best average rank (\textbf{1.625}), but their characteristics are different. BiSWE remains competitive in overall stability, while MMCI achieves the highest global Spearman (\textbf{0.9363}) and the largest adjacent-level separation (\textbf{3.2218}), indicating stronger discriminative ability under complex mixed perturbations. In contrast, MS2R is consistently weaker across all three settings, especially under \emph{mixed hard}, where its global correlation becomes near-zero and its monotonic rate drops to $0.0$, showing limited robustness under severe temporal distortions.

Overall, these results verify that the proposed MMCI and TCPE are not redundant with existing temporal metrics, but instead provide more task-aligned and interpretable temporal diagnostics. Specifically, MMCI is particularly effective for characterizing \emph{cross-frame modal-mixing continuity}, while TCPE is especially suitable for measuring \emph{temporal source-correlation preservation}. This validates the effectiveness of our metric design and further supports the novelty of introducing more fine-grained temporal evaluation for infrared-visible video fusion.

\section{Analysis of Controlled Perturbations on Low-/High-Frequency Branches}
\label{app:lfhf_analysis}

To further examine whether the proposed frequency-aware design is consistent with the target perturbations considered in the main paper, we analyze how three controlled degradations, i.e., frame-wise flicker, temporal jitter, and local misalignment, are distributed after the low-/high-frequency decomposition. Given an input sequence $\{F_t\}_{t=1}^{T}$, we compute its low-frequency and high-frequency components by
\begin{equation}
L_t, H_t = \mathcal{D}(F_t),
\end{equation}
and measure their temporal variation energies as
\begin{equation}
E_L = \frac{1}{T-1}\sum_{t=2}^{T}\|L_t-L_{t-1}\|_2^2,\qquad
E_H = \frac{1}{T-1}\sum_{t=2}^{T}\|H_t-H_{t-1}\|_2^2.
\end{equation}
We further define the low-frequency energy ratio as
\begin{equation}
R_L = \frac{E_L}{E_L+E_H},
\end{equation}
and report the temporal low-band energy ratios of the two branches as $\eta_L$ and $\eta_H$, respectively.

The results are summarized in Table~\ref{tab:lfhf_controlled_analysis}. Under \emph{flicker}, $E_L$ increases from $7.34\times 10^{-4}$ to $9.61\times 10^{-4}$, while $E_H$ remains almost unchanged, causing $R_L$ to rise from $0.5898$ to $0.6910$. This indicates that flicker-induced temporal variation is mainly absorbed by the low-frequency branch. Under \emph{local misalignment}, the effect is even more pronounced: $R_L$ increases from $0.5898$ to $0.8562$, showing that the resulting cross-frame variation is also dominated by the low-frequency component. In contrast, under \emph{jitter}, $R_L$ gradually decreases from $0.5898$ to $0.5024$, suggesting that this perturbation is distributed more evenly across the two branches under the current decomposition. Nevertheless, $\eta_L$ remains consistently higher than $\eta_H$ for all three perturbations, indicating that the low-frequency branch still preserves more slow-varying temporal content.

Overall, these observations support the proposed strategy from two aspects. First, the low-frequency branch is indeed the primary carrier of slow temporal disturbances such as flicker and local misalignment, which justifies performing temporal stabilization and shared temporal context modeling in this branch. Second, the more balanced LF/HF response under jitter suggests that not all temporal perturbations should be treated as purely low-frequency effects, which further motivates the complementary high-frequency modeling introduced by SCAM.

\begin{table}[t]
\centering
\caption{Temporal energy analysis under controlled perturbations. $E_L$ and $E_H$ denote the temporal variation energies of the low- and high-frequency branches, respectively; $R_L$ is the low-frequency energy ratio; $\eta_L$ and $\eta_H$ are the temporal low-band energy ratios of the two branches.}
\label{tab:lfhf_controlled_analysis}
\resizebox{0.98\linewidth}{!}{
\begin{tabular}{c|c|ccccc}
\hline
Perturbation & Strength & $E_L$ & $E_H$ & $R_L$ & $\eta_L$ & $\eta_H$ \\
\hline
\multirow{5}{*}{Flicker}
& 0.0 & 0.000734 & 0.000371 & 0.589846 & 0.996621 & 0.940629 \\
& 0.5 & 0.000748 & 0.000371 & 0.598594 & 0.996591 & 0.940627 \\
& 1.0 & 0.000793 & 0.000371 & 0.616832 & 0.996481 & 0.940639 \\
& 1.5 & 0.000864 & 0.000371 & 0.659099 & 0.996323 & 0.940599 \\
& 2.0 & 0.000961 & 0.000371 & 0.691016 & 0.996096 & 0.940567 \\
\hline
\multirow{5}{*}{Jitter}
& 0.00 & 0.000734 & 0.000371 & 0.589846 & 0.996621 & 0.940629 \\
& 0.15 & 0.000587 & 0.000355 & 0.565292 & 0.997191 & 0.943540 \\
& 0.25 & 0.000504 & 0.000346 & 0.548290 & 0.997521 & 0.945350 \\
& 0.40 & 0.000403 & 0.000332 & 0.523369 & 0.997942 & 0.947864 \\
& 0.55 & 0.000330 & 0.000320 & 0.502409 & 0.998274 & 0.950127 \\
\hline
\multirow{5}{*}{Local Misal.}
& 0.0 & 0.000734 & 0.000371 & 0.589846 & 0.996621 & 0.940629 \\
& 1.0 & 0.001123 & 0.000428 & 0.697222 & 0.995764 & 0.934968 \\
& 2.0 & 0.001743 & 0.000494 & 0.769398 & 0.994278 & 0.926875 \\
& 3.0 & 0.002676 & 0.000560 & 0.828635 & 0.992190 & 0.919818 \\
& 4.0 & 0.003563 & 0.000596 & 0.856231 & 0.990245 & 0.914410 \\
\hline
\end{tabular}}
\end{table}

\begin{figure*}[t]
  \centering
  \begin{subfigure}[t]{0.48\linewidth}
    \centering
    \includegraphics[width=\linewidth]{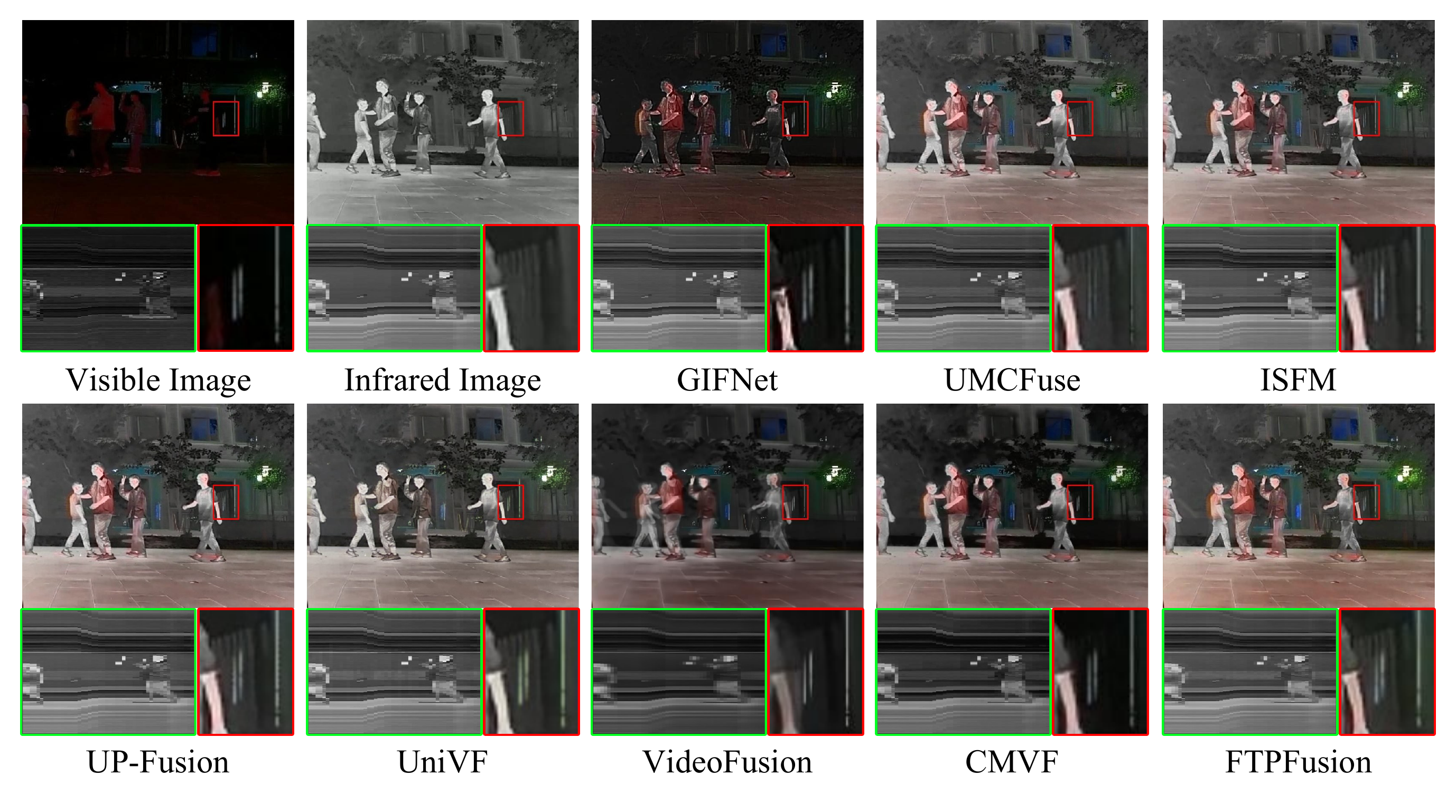}
    \caption{Qualitative comparison on the HDO dataset.}
    \label{figA2}
  \end{subfigure}
  \hfill
  \begin{subfigure}[t]{0.48\linewidth}
    \centering
    \includegraphics[width=\linewidth]{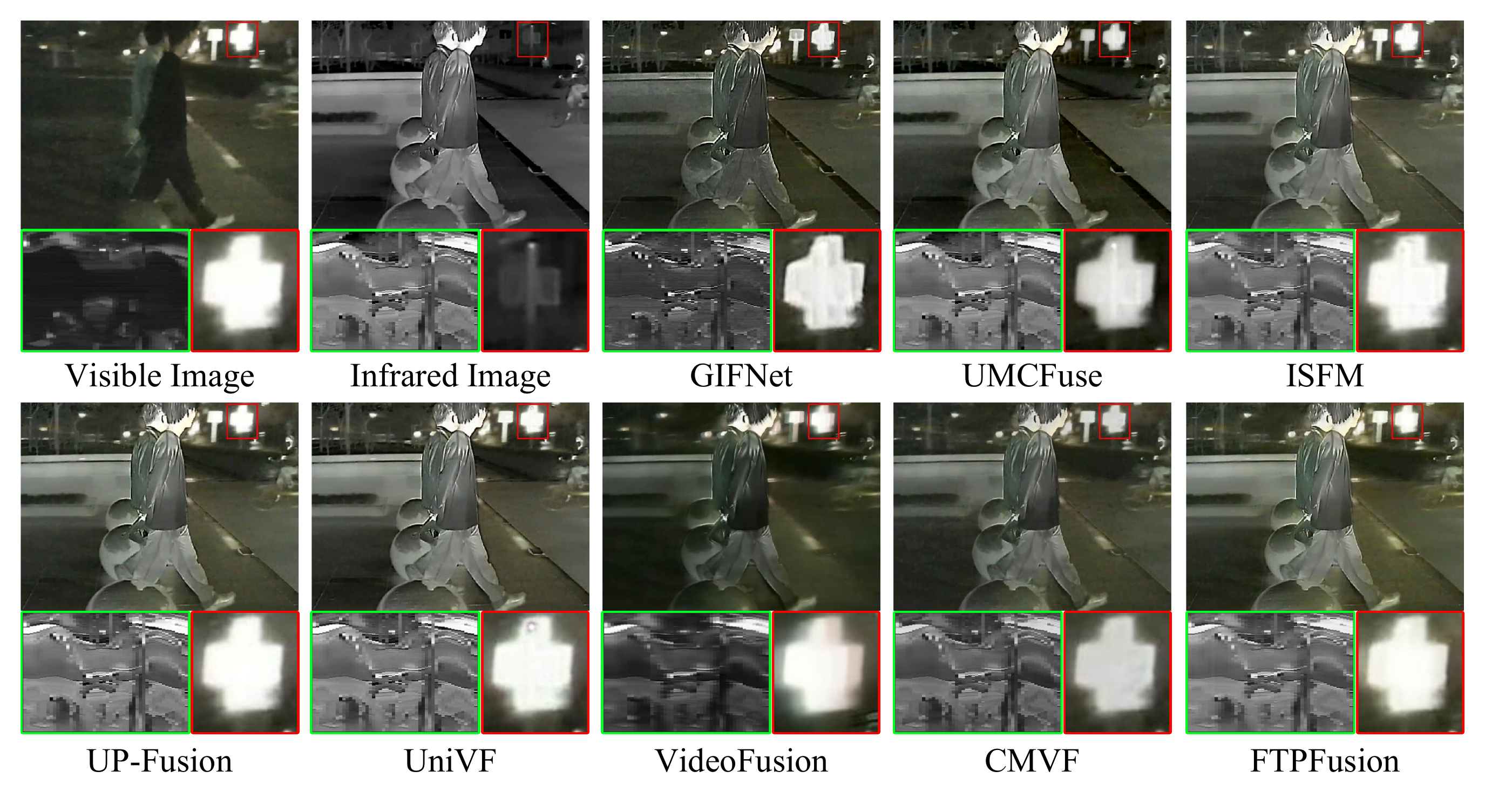}
    \caption{Qualitative comparison on the VTMOT dataset.}
    \label{figA3}
  \end{subfigure}
  \caption{Qualitative comparison of all methods on two datasets.}
  \label{fig:qualitative_compare}
\end{figure*}

\section{Fusion Performance Comparison under Normal Scenes}

\begin{figure}[t]
  \centering
   \includegraphics[width=1\linewidth]{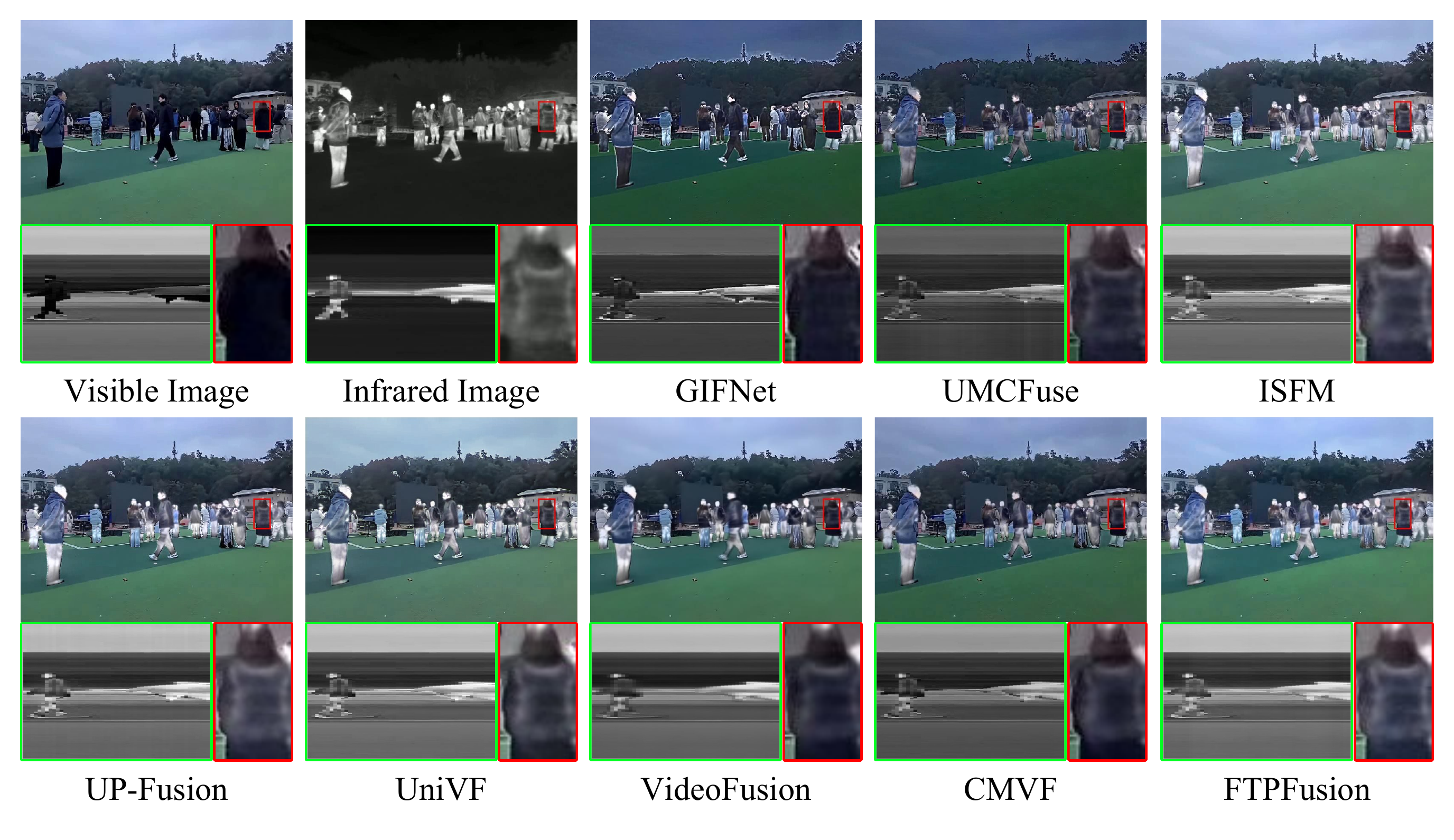}
   \caption{Qualitative comparison of all methods on the M3SVD dataset.}
   \label{figA1}
\end{figure}

\Cref{figA1,figA2,figA3} present supplemental qualitative comparisons on M3SVD, HDO, and VTMOT under normal scenes. Red boxes highlight key local regions, while green boxes denote temporal slices. These slices are constructed by stacking pixels at a fixed column position over time to visualize brightness variations and structural continuity. Generally, existing methods exhibit a clear trade-off between spatial detail preservation and temporal consistency. Image fusion methods such as ISFM and UP-Fusion preserve certain brightness and texture but remain insufficient in local structure and temporal stability, often leading to blurriness. Conversely, video fusion methods show better temporal continuity but suffer from various degrees of detail degradation. Specifically, UniVF favors spatial detail representation, while VideoFusion provides better temporal stability. In comparison, our method preserves object contours and local textures more completely across all datasets. The corresponding temporal slices exhibit smoother and more natural structural transitions, which indicates that our approach effectively mitigates inter-frame brightness jumps and local jitter. These results further demonstrate that while existing methods often prioritize either spatial recovery or temporal modeling, our method achieves a superior balance with stronger robustness and generalization.

\end{document}